%% file: example_paper.tex
\documentclass{article}

\usepackage{microtype}
\usepackage{graphicx}
\usepackage{subfigure}
\usepackage{booktabs} %
\usepackage{amsmath, amssymb}
\usepackage{amsfonts}
\usepackage{listings}
\usepackage{xcolor}
\usepackage{soul}
\usepackage{tikz}
\usepackage{pifont}
\usepackage{caption}
\usepackage[shortlabels]{enumitem}
\usepackage{algpseudocode}
\usepackage{listings}
\usepackage{enumitem}

\usepackage{xcolor}

\usepackage{hyperref}
\usepackage{cleveref}

\algrenewcommand\algorithmicrequire{\textbf{Input:}}
\algrenewcommand\algorithmicensure{\textbf{Output:}}
\usepackage{mlsys2024}
\mlsystitlerunning{Self-Selected Attention Span for Accelerating Large Language Model Inference}

\begin{document}

\twocolumn[
    \mlsystitle{Self-Selected Attention Span \\ for Accelerating Large Language Model Inference}

    \mlsyssetsymbol{equal}{*}

    \begin{mlsysauthorlist}
        \mlsysauthor{Tian Jin}{dm,mit}
        \mlsysauthor{Wanzin Yazar}{dm}
        \mlsysauthor{Zifei Xu}{dm}
        \mlsysauthor{Sayeh Sharify}{dm}
        \mlsysauthor{Xin Wang}{dm}
    \end{mlsysauthorlist}

    \mlsysaffiliation{dm}{d-Matrix Corporation, Santa Clara, California, USA}
    \mlsysaffiliation{mit}{MIT, Cambridge, Massachusetts, USA}

    \mlsyscorrespondingauthor{Tian Jin}{tianjin@csail.mit.edu}
    \mlsyscorrespondingauthor{Xin Wang}{xwang@d-matrix.ai}

    \mlsyskeywords{Machine Learning, MLSys}

    \vskip 0.3in

    \input{abs.tex}
]

\printAffiliationsAndNotice{\mlsysEqualContribution} %

\input{intro.tex}
\input{bg.tex}
\input{meth.tex}

\input{exp_res.tex}

\input{related.tex}

\section{Closing Discussions}

\textbf{Interpretability.}
Large language models can cost up to millions of dollars to train \citep{Venigalla_Li}.
This high cost makes developers of LLMs rightfully hesitant to use opaque model compression techniques that might compromise model quality.
We believe that, in order for such techniques to gain widespread adoption, model developers must be able to see and understand how the model context is reduced and compressed.
Our method meets this criterion by reducing the context size of these models in fully transparent ways.
We train LLMs to produce easy-to-inspect attention span annotations that appear in the model output.
These annotations help developers determine whether the context reduction is justified.

\textbf{Debugging and intervention.}
Interpretable attention sparsity facilitates debugging.
If model quality falls below an acceptable threshold, model developers can assess the selected attention span by inspecting the anchors and references in the model's output.
Finally, if developers determine that the selections are incorrect, they can correct the attention span annotations by creating a dataset with the correct annotations and subsequently fine-tuning the model.

\textbf{Conclusion.}
Our work shows that fine-tuning LLMs to predict and minimize its own attention spans can accelerate autoregressive inference by up to 28\%.
We label task datasets with attention span annotations, design custom CUDA kernels to improve inference efficiency.
We believe our study is a step towards making LLMs autonomously optimize its own computations.

\bibliography{example_paper}
\bibliographystyle{mlsys2024}

\makeatletter
\newcommand{\crefnames}[3]{%
    \@for\next:=#1\do{%
        \expandafter\crefname\expandafter{\next}{#2}{#3}%
    }%
}
\makeatother

\crefnames{part,chapter,section}{\S}{\S\S}

\appendix
\section{Example prompts used for generating annotated summarization dataset}
\label{sec:prompt}
Prompt used for ChatGPT to generate summary:
\begin{quote}
\textsl{Provide a summary of the above article in no more than three sentences. 
For each sentence, precede it with the range of line numbers from the article that informed it, using the format (line numbers). For example: (1-3) This is the first sentence. Write each sentence as a separate sentence on a new line.}
\end{quote}
Prompt attached to instructions in the new dataset:
\begin{quote}
\textsl{Summarize the following article in three sentences and precede it with the range of line numbers for each summarized sentence in the parenthese.}
\end{quote}

\end{document}

%% file: abs.tex
\begin{abstract}
    Large language models (LLMs) can solve challenging tasks.
    However, their inference computation on modern GPUs is highly inefficient due to the increasing number of tokens they must attend to as they generate new ones.
    To address this inefficiency, we capitalize on LLMs' problem-solving capabilities to optimize their own inference-time efficiency.
    We demonstrate with two specific tasks: (a) evaluating complex arithmetic expressions and (b) summarizing news articles.
    For both tasks, we create custom datasets to fine-tune an LLM.
    The goal of fine-tuning is twofold: first, to make the LLM learn to solve the evaluation or summarization task, and second, to train it to identify the minimal attention spans required for each step of the task.
    As a result, the fine-tuned model is able to convert these self-identified minimal attention spans into sparse attention masks on-the-fly during inference.
    We develop a custom CUDA kernel to take advantage of the reduced context to attend to.
    We demonstrate that using this custom CUDA kernel improves the throughput of LLM inference by $28$\%.
    Our work presents an end-to-end demonstration showing that training LLMs to self-select their attention spans speeds up autoregressive inference in solving real-world tasks. 
\end{abstract}

%% file: intro.tex
\section{Introduction}

Recent advances of large language models (LLMs) showed impressive capabilities \citep{devlin2019bert, brown2020language, wei2022emergent, chowdhery2022palm, touvron2023llama, zhang2022opt, biderman2023pythia, chen2021evaluating}.
However, the computation required to perform inference with LLMs is highly inefficient on state-of-the-art hardware.
This inefficiency presents a pressing obstacle to sustainable and widespread deployment of LLMs.

The autoregressive nature of LLM inference underpins the low arithmetic intensity during inference.
Generating every new token depends on all preceding tokens as its context.
This dependency necessitates storing and reading a growing size of contextual states, leading to suboptimal GPU utilization and memory-related performance bottlenecks.

To speed up the attention mechanism, 
existing techniques either adopt a static sparsity patterns in the attention matrix~\citep{beltagy2020longformer, zaheer2021big} or introduce external mechanisms that dynamically induce attention sparsity~\citep{tay2020sparse, wang2021spatten, rao2021dynamicvit}.
The latter methods all employ a \emph{context sparsifier}, an add-on module external to the LLM, co-optimized to reduce the context that the attention mechanism attends to. 
The separation between this context sparsifier module from the LLM itself prevents them from capitalizing on the ever-growing capabilities of LLMs themselves to sparsify attention.

Meanwhile, recent LLMs excel at zero-shot and few-shot learning tasks, showing generalization abilities across a wide variety of tasks, leading to the hypothesis of genuine reasoning ability and general intelligence. 
In this work, we hypothesize and validate that LLMs themselves possess the capability to serve as the aforementioned context sparsifier.  

\begin{figure}[t!]
    \centering
    \includegraphics[width=0.95\linewidth]{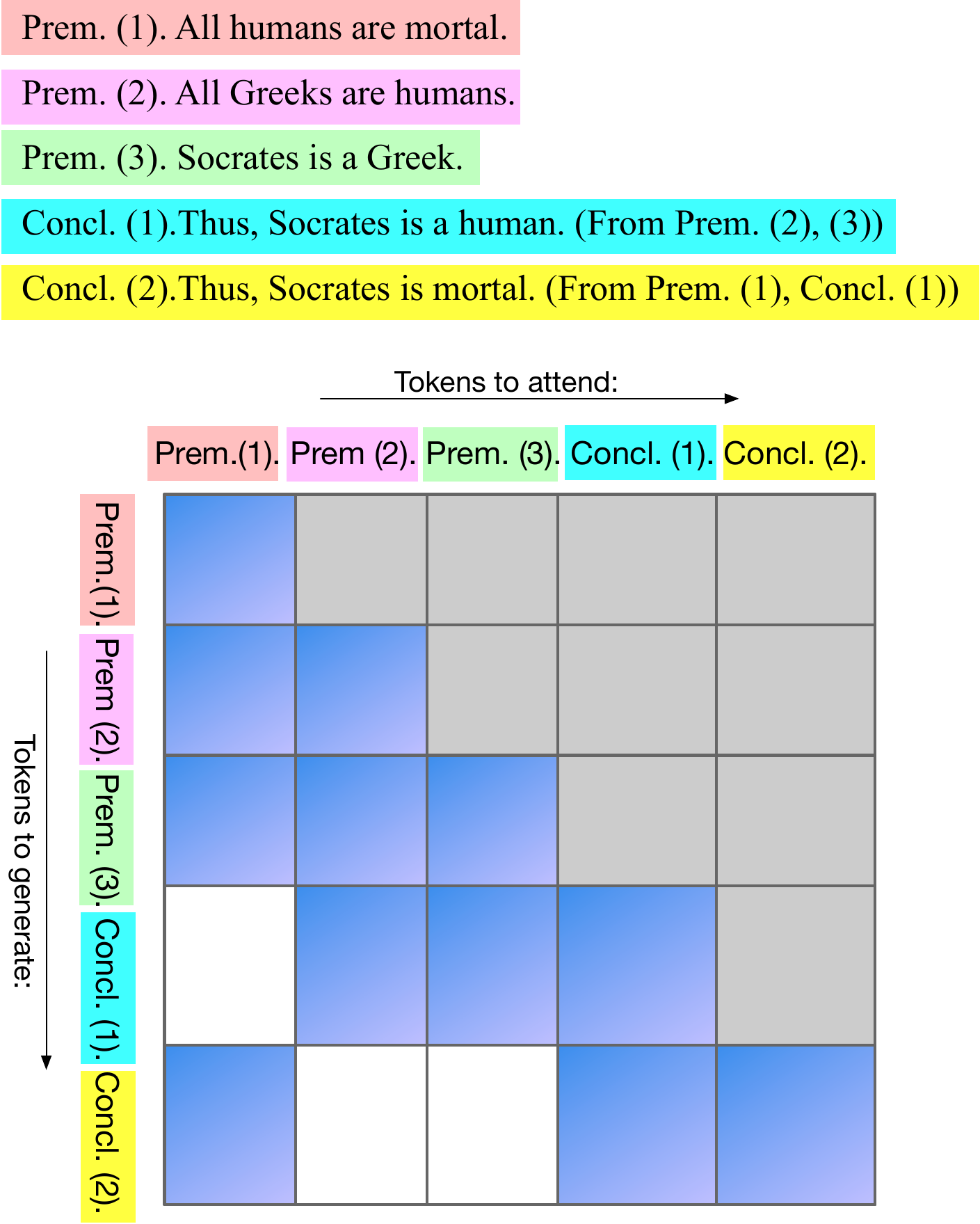}
    \caption{Human thought process is inherently sparse, as shown by the minimal dependencies in the attention matrix below.}
    \label{fig:intro}
\end{figure}

To illustrate our approach, consider the chained syllogisms in \Cref{fig:intro}, an example of inherent sparsity in logical reasoning.
Concluding that Socrates is Greek needs only the second and the third premise.
Similarly, concluding that Socrates is mortal needs only the first premise and the first conclusion.
Our work uses LLMs themselves to predict such natural, coarse-grained attention sparsity.
We design custom CUDA kernels to speed up inference on GPUs. 

\paragraph{Contributions:}
\begin{enumerate}[leftmargin=1.2em]
\itemsep0em 
\item We propose using large language models to self-select important subsets of tokens (referred to as \textit{attention span}) in its context to improve its own inference-time efficiency. %

\item We create two annotated datasets for complex arithmetic expression evaluation and for news article summarization to fine-tune large language models. 
These datasets include not only the solution to the task, but annotations of the minimal attention span required to solve the task.

\item We analyze and benchmark key components of state-of-the-art LLMs, explaining how optimizing individual components affects full-model performance.

\item We design a custom GPU kernel to accelerate autoregressive inference with self-selected attention spans and achieve up to 28\% inference throughput increase
without compromising accuracy on arithmetic evaluation task.

\item We analyze how attention sparsity evolves over time in our approach, and document how LLMs make dynamic and context-dependent prediction of the required attention span for generating next tokens. 

\item We benchmark our kernel implementations and identify conditions under which our custom CUDA kernels speed up LLM inference.

\end{enumerate}

\paragraph{Implications.}
Our work demonstrates the benefits of using insights from LLMs for improving thier own inference efficiency.
Our work constitutes a preliminary step towards having LLMs autonomously and adaptively optimize their own computation at inference-time. 

%% file: bg.tex
\section{Background}
\label{sec:background}

In this section, we describe important model architecture details (\Cref{subsec:model-arch}), explain their implications on model runtime (\Cref{subsec:autoregressive-inference}) and breakdown model runtime to individual components (\Cref{subsec:runtime-breakdown}).

\subsection{Model architecture}
\label{subsec:model-arch}

We consider state-of-the-art decoder-only LLMs \citep{zhang2022opt, touvron2023llama, pmlr-v202-biderman23a}.
The architecture of these models commonly features repeated instances of a building block called the \textit{decoder}.
In the range of models we consider, each decoder consists of a self-attention layer succeeded by a feedforward network.
Additionally, the architecture includes normalization and positional embedding layers, which have inconsequential impact on overall system performance.

\subsection{Autoregressive inference}
\label{subsec:autoregressive-inference}
To generate content using a LLM, one starts with a prompt, consisting of $l$ initial tokens $X = (p_0, p_1,\cdots, p_{l-1}) \in \mathbb{Z}^{l}$.
The model sequentially generates output tokens.
At each step $t$, the model computes the probability of the next token $o_{l+t}$ conditioned on the concatenation of prompt tokens $(p_0, p_1,\cdots, p_{l-1})$ and previously generated output tokens $(o_{l},\cdots,o_{l+t-1})$: $P(o_{l+t}|p_0,\cdots, p_{l-1}, o_{l},\cdots,o_{l+t-1}).$
Next, we explain the computationally expensive steps involved in computing this conditional probability.

\paragraph{Self-attention.} Suppose the input to the decoder module consists of $d$-dimensional vectors corresponding to $l$ tokens: $X = (x_0, x_1,\cdots, x_{l-1}) \in \mathbb{R}^{l \times d}$.
Self-attention computes matrices for queries \( Q \), keys \( K \), and values \( V \) as:

\begin{align}
    Q & = XW_Q \in \mathbb{R}^{l \times d} \\
    K & = XW_K \in \mathbb{R}^{l \times d} \\
    V & = XW_V \in \mathbb{R}^{l \times d}
\end{align}

Next, the attention score matrix \( S \), is computed as:

\begin{equation}
    S = \frac{QK^T}{\sqrt{d}} \in \mathbb{R}^{l \times l}
\end{equation}

An \textit{attention mask} $M$ is then added to $S$.
An attention mask is a matrix that determines the data dependency between tokens -- element at position [i, j] determines whether the generation of token at position $i$ can attend to token at position $j$.
A value of $0$ in M permits attention, while a value of negative infinity prohibits it.
We obtain the attention probability matrix \( A \), by applying the softmax function row-wise to $M \oplus S$.
The softmax function ignores positions with large negative values in the mask, and focus only on those positions with zeros in the mask.

\begin{equation}
    A_{ij} = \text{Softmax}_j(M_{i, :} \oplus S_{i, :})
\end{equation}

Finally, the output for each token after the self-attention operation is given by:

\begin{equation}
    O = AV \in \mathbb{R}^{l \times d}
\end{equation}

Each row \( O_i \) of \( O \) correspond to an input token \( x_i \) after the self-attention operation.

\paragraph{Feedforward network.}
The feedforward network compute a series of linear projections for each output vector $o_i$, interspersed with activation functions between them.

\paragraph{KV-cache.}
Since autoregressive inference generates output tokens sequentially, at each token generation step, key and value vectors associated with prior tokens remain unchanged across generation steps.
Specifically, for the $t$-th output token generation, the key and value vectors corresponding to tokens $x_i$ with $i < t$ remain unchanged from previous computations.

This observation leads to the introduction of a \textit{key-value cache (KV-cache)}: once we compute the key and value vectors for a token, they are cached in memory.
In subsequent generation steps, rather than recomputing these vectors, we retrieve them from the cache.
KV-cache significantly reduces the computational cost of token generation steps.

\paragraph{Token generation.}
We refer to the generation of the first token as the \textit{prefill} phase of the inference.
KV-cache does not change the computational complexity of prefill.
This step computes queries, keys, values, and attention scores/outputs for all the prompt tokens in parallel using batched matrix multiplication.
Optimized kernels exist to execute them efficiently on state-of-the-art hardwares ~\citep{chen2018tvm, 10.1145/3315508.3329973}.

We refer to the generation of subsequent tokens as the \textit{decoding} phase of the inference.
As LLM generates more tokens, decoding becomes increasingly memory-intensive.
Generation of token $o_{l+t}$ only requires the computation of $(l+t)$-th row of $O$. 
This corresponds to calculating a single row of the attention score and the attention probability matrices, $S_{l+t}$ and $A_{l+t}$, 
which necessitates reading just one row from the query matrix, $Q$. However, this is not the case with the key and value matrices. Since the self-attention mechanism considers the relationship of the current token with all the previously generated tokens, reading the complete matrices $K$ and $V$ is essential to computing the single row of attention score $S_{l+t}$ 
corresponding to the current token.

Therefore, the calculations for subsequent token generations largely rely on matrix-vector operations.
As these operations are inherently memory-bound, optimizing memory access—specifically by reducing memory reads— can speedup token generation.

\subsection{Performance characterization}
\label{subsec:runtime-breakdown}

\paragraph{Runtime breakdown.}
\begin{figure}[h]
    \includegraphics[width=.5\textwidth, trim={2cm 0 1.5cm 1cm},clip]{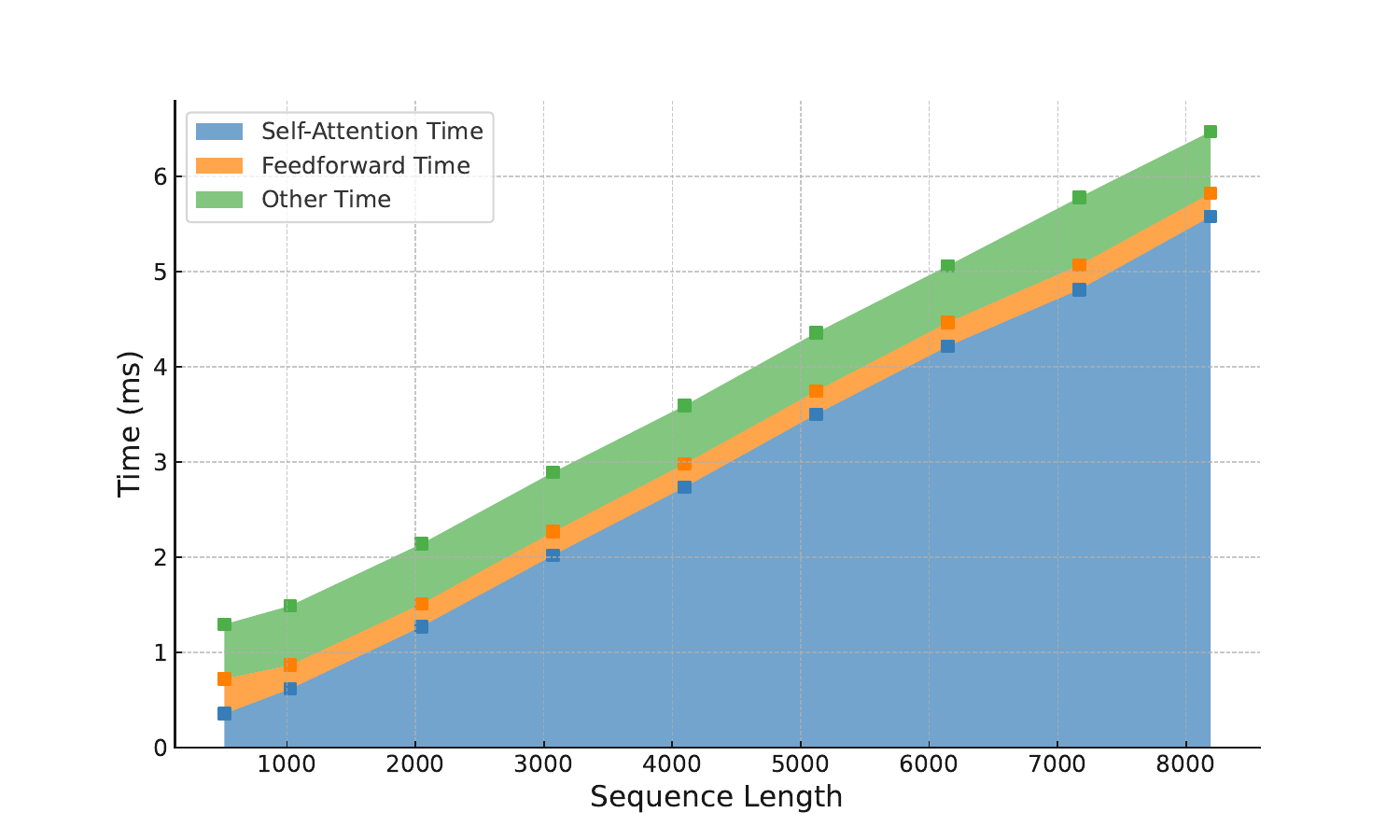}
    \caption{Runtime breakdown for decoder execution during inference by LLaMA-7B model on an A100 GPU. At $2048$ sequence length, $50$\% attention sparsity yields a maximum runtime reduction of $29.7$\%, assuming linear speedup w.r.t. sparsity..}
    \label{fig:runtime-breakdown}
\end{figure}

We investigate language models consisting of repeated decoder blocks that possess identical architectures.
Using LLaMA-7B~\citep{touvron2023llama} as a representative model, we benchmark a single decoder layer.
We analyze and break down the time required to produce a single output token using the Hugging Face transformers~\citep{wolf-etal-2020-transformers} implementation of the LLaMA model.
We experiment with a batch size of $64$, the largest batch size to fit all sequence lengths we investigate,
and vary the number of preceding tokens to attend to, which include both the prompt and previously generated tokens, on an A100 GPU, with KV-cache enabled and pre-allocated. 

\Cref{fig:runtime-breakdown} shows that as the sequence length increases to $512$, $1024$, and $2048$ tokens, self-attention consumes $27.4$\%, $41.4$\%, and $59.4$\% of the total computation time, respectively.  
Thus, optimizing self-attention could theoretically reduce computation time by up to $29.7$\% with $50$\% sparsity when generating sequences shorter than 2048 tokens.
The corresponding percentage for feedforward network is $28.3$\%, $16.8$\% and $11.2$\%, correspondingly.

%% file: meth.tex
\section{Self-selected attention span}

\input{code/method-example}

As \Cref{sec:background} explains, reducing memory reads during self-attention can speed up autoregressive inference by LLMs.
We propose to reduce the attention span at each token generation step; specifically, limiting the number of tokens each output generation step considers, thereby reducing memory reads during self-attention computation.

\subsection{Motivation}
Sequential arithmetic evaluation is a typical task where generation of current token does not depend on the entire context. 
Take the task of evaluating simple arithmetic expression ``$10 + 42 \times 3$'' for example.
Prompting the model with tokens ``$10 + 42 \times 3 =$'' may result in the following sequence of reasoning steps and answer:

\vspace{-2em}
\begin{center}
    \notag
    \begin{align}
         & 10 + 42 \times 3 \\
         & = 10 + 126       \\
         & = 136
    \end{align}
\end{center}
\vspace{-1em}

For a model trained to evaluate arithmetic expressions, each generation step attends to all preceding tokens, including the initial prompt and all earlier reasoning steps.
However, by logic, to generate the final line, the model only needs the preceding reasoning step.
Distant historical context like the initial prompt should not, in principle, influence the quality of the final answer.
This illustrates potential for narrowing the model's attention span.
Notably, similar patterns of redundancy exists broadly in tasks such as summarization, multi-round question-answering, and other tasks requiring chain-of-thought reasoning.

\subsection{Inference}

We introduce our approach for autoregressive inference with a reduced number of tokens to attend to. %

\textbf{Challenges.}
Correctly identifying the necessary attention span to generate a token is challenging.
In tasks that entail constructing reasoning steps with complex logical dependencies, identifying the minimal attention span equates to pinpointing the logical dependencies of the upcoming reasoning step.
Due to its challenging nature, we hypothesize that the LLM itself is best suited to perform this step.
To facilitate this process, we use the following design.

\textbf{Context partitioning.}
We partition the context of an LLM into semantically meaningful groups of contiguous tokens.
This step can often benefit from human design -- for arithmetic expression evaluation task, tokens corresponding to a single step of calculation can naturally be grouped together. For summarization task, tokens within the same sentence will be grouped together.

\textbf{Anchors.}
We introduce the concept of \textit{anchors} to identify each group of tokens.
An anchor consists of one or more tokens, and follows the group of tokens it identifies.
We can manually insert these anchors into the input prompt or generate them alongside other output tokens by the LLM trained to auto-partition its output as discussed in the previous step.

\textbf{Determining attention span.}
Next, we use an LLM to determine the necessary attention span for the next token prediction.
In this step, we allow LLMs to attend to all preceding context. 
Offline, we fine-tune the LLM with datasets designed to demonstrate attention span identification.
Creating these datasets is task-specific, and we offer examples of how to make them in our experimental evaluations.

\textbf{Token generation.}
For output token generations, we restrict the LLMs to attend only to tokens corresponding to the attention span denoted by these anchors.

Here, we use the same example as illustration in \Cref{fig:example}.
\begin{enumerate}[(a), leftmargin=0.6cm]
\itemsep0em 
    \item We start with a prompt containing input tokens for the arithmetic expression $10 + 42 \times 3$. 
    We manually add an anchor (\circled{1}) to the prompt to mark the initial step.
    \item The LLM identifies the smallest attention span necessary to correctly produce the next calculation step. The model chooses anchor \circled{1}, signifying the need for the first calculation step to produce subsequent tokens.
    \item The LLM accurately produces the second step of calculation, along with a new anchor \circled{2}.
    \item At this step, the model selects anchor \circled{2} as its attention span. It ignores the tokens spanned by \circled{1} as \circled{2} encompasses the information of \circled{1}.
    \item Finally, the model generates the correct answer, attending solely on the preceding step.
\end{enumerate}

\textbf{Overhead.} 
Our proposed method, while enhancing inference efficiency, incurs a minor runtime overhead due to the generation of anchors and references. 
We emphasize that this overhead is minimal.\footnote{In \Cref{custom-cuda-kernel-benchmark}, we measure this overhead to be $\le7\%$.}
The LLM generates anchors and references for groups of tokens—for example, those associated with the same evaluation step—rather than on a per-token basis. 
Therefore, we can amortize this overhead across the generation of numerous tokens within each group, significantly mitigating its overall impact.

\begin{figure*}[t]
    \includegraphics[width=1\textwidth]{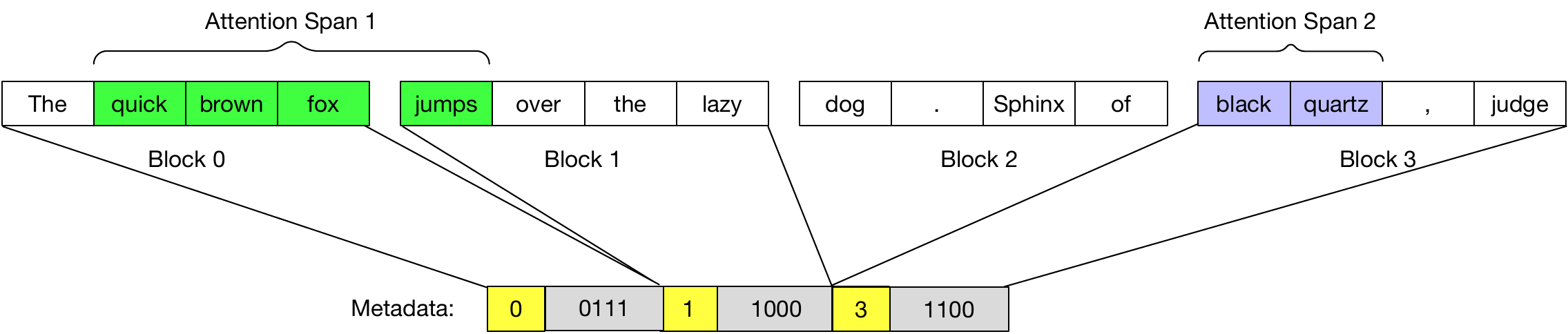}
    \caption{Illustration of attention span encoding. Numbers on a yellow background are block indices. Numbers on gray background represent the binary mask indicating whether each token within the block should be attended to.}
    \label{fig:attn-span-encoding}
\end{figure*}

\subsection{CUDA Kernel Design}

The proposed approach reduces a LLM's context during token generation to decrease memory reads during inference.
However, efficiently implementing this approach is challenging.
Specifically, the identified attention span for the next token generation might consists of non-contiguous token spans.
Optimized self-attention implementations typically assume contiguous input tokens \citep{dao2022flashattention}.

A straightforward solution is to repack these non-contiguous attention spans into a single contiguous segment, excluding the irrelevant context.
This approach, however, faces two issues: Firstly, it necessitates copying the key and value matrices every time we switch between attention span determination and token generation.
These repeated memory operations further stress the memory bandwidth during memory-bound computations, leading to slower inference speed.
Secondly, the memory associated with the full context cannot be freed since subsequent token generations might reference previously unused contexts.
This could require maintaining two context versions, straining the limited GPU memory capacity.

\textbf{Attention span encoding.}
We introduce a blocked attention span encoding method. First, we divide the context tokens into blocks of a fixed size, e.g. $256$ tokens each.
Within each block, we record two metadata components: its block index, and a block mask recording which tokens in the block should be attended to.
We omit recording metadata for blocks that do not contain any elements of the selected attention span.
When encoding attention spans for multiple requests within a single batch, we maintain separate metadata for each request.
We illustrate this scheme in \Cref{fig:attn-span-encoding}.

\begin{algorithm}[b!]
    \begin{algorithmic}[1]
        \Require $d$ (Hidden dimension size)
        \Require $n$ (Number of tokens in current context)
        \Require $bs$ (Block size)
        \Require $q$ (Query vector of dimension $d$)
        \Require $K$ (Key matrix of size $[n \times d]$)
        \Require $V$ (Value matrix of size $[n \times d]$)
        \Require $metadata$ (List of tuples containing block indices and start/end indices within blocks)
        \Ensure $o$ (Output vector after attention computation)

        \Function{compute\_attn}{$q$, $K$, $V$, $metadata$}
        \State $qK, o \gets \text{fill}(0, [n]), \text{fill}(0, [n])$
        \For{each $(block\_idx, bm)$ in $metadata$}
        \State $bm = (bm - 1) \times -inf$
        \State $\delta \gets block\_idx \times bs$
        \State $bqK \gets \text{fill}(0, [bs])$
        \State \textbf{parallel for} $i$ from $0$ to $bs$
        \State \hspace{.6cm}$bqK[i] \gets \text{dot}(q, K[\delta + i, :])$
        \State \textbf{end parallel for}
        \State $qK[\delta:\delta+bs] = bqK \oplus bm$
        \EndFor
        \State $A \gets \text{softmax}(qK/\sqrt{d})$
        \For{each $(block\_idx, start, end)$ in $metadata$}
        \State $\delta \gets block\_idx \times bs$
        \State \textbf{parallel for} $i$ from $0$ to $bs$
        \State \hspace{.6cm}$o[\delta + i] \gets A[\delta + i] \times V[\delta + i, :]$
        \State \textbf{end parallel for}
        \EndFor

        \State \Return $o$
        \EndFunction
    \end{algorithmic}
    \caption{Computing the self-attention during autoregressive generation for a single output token.}
    \label{alg:sparse-attn}
\end{algorithm}

\textbf{Kernel algorithm.}
We modify the FlashAttention~\citep{dao2022flashattention} kernel implemented in the OpenAI Triton project~\citep{10.1145/3315508.3329973}.
We show our modified algorithm in \Cref{alg:sparse-attn}.
We dispatch an instance (CUDA block) of the algorithm for each attention head and batch of input.

In L3-11 of \Cref{alg:sparse-attn}, we calculate $qK^T$, i.e. the dot product of the query vector with each row of the key matrix, for every block specified in the metadata.
L4 converts the block mask from a binary mask to a softmax mask. This mask ensures attention probabilities are non-zero only for positions corresponding to non-zero element in the block mask.
L5 determines the block's starting offset, $\delta$.
L6 initialize $bqK$, the block-level representation of $qK^T$, to zero..
L7-9 use parallel threads in a CUDA threadblock to compute the dot product for each token in the block.
In L10, block mask $bm$ ensures only tokens within the selected attention span influence the attention probabilities, $A$.
Lastly, L13-18 calculate the attention output using attention probabilities $A$ and the value matrix $V$, again using parallel threads in a CUDA threadblock.

Overall, to produce each output token, we compute self-attention only over tokens inside the selected attention span.
Blocks without tokens from the selected span are excluded from attention computation, thereby reducing memory reads and computation.
For blocks partially intersecting the attention span, we mask out tokens that fall outside, ensuring they don't contribute to attention probabilities.
Maximizing the number of blocks without attention span overlap is essential to accelerate inference.

\textbf{Auto-tuning.} We use Triton autotuner to select optimal block size and thread count per CUDA block.

%% file: code/method-example.tex
\newcommand{\circled}[1]{\tikz[baseline=(char.base)]{
            \node[shape=circle,draw,inner sep=0.5pt] (char) {#1};}}

\lstnewenvironment{teX}[1][]
{
    \lstset{language=[LaTeX]TeX}
    \lstset{
        escapeinside={(*@}{@*)},
        breaklines=true,
        basicstyle=\normalsize\ttfamily,
        showstringspaces=false,
        keywordstyle=\itshape\color{blue},
        stringstyle=\color{maroon},
        commentstyle=\color{black},
        rulecolor=\color{black},
        xleftmargin=0pt,
        xrightmargin=0pt,
        aboveskip=\medskipamount,
        belowskip=\medskipamount,
        backgroundcolor=\color{white},
        #1
    }
}
{}

\newcommand{\highlightandunderline}[1]{%
    \tikz[baseline=(A.base)] \node[fill=green, draw=none, inner sep=1pt] (A) {\ul{#1}};%
}

\sethlcolor{green}

\begin{figure*}
\centering

\begin{minipage}[t]{0.18\linewidth}
\captionsetup[figure]{labelfont={bf},labelformat={default},labelsep=period,name={Fig.}}
    \begin{teX}
10 + 42 * 3=(*@\circled{1}@*)
    \end{teX}
(a) Input Prompt
\end{minipage}%
\begin{minipage}[t]{0.21\linewidth}
    \captionsetup{labelformat=empty, labelsep=space}
    \begin{teX}
(*@ \hl{10 + 42 * 3=}@*)(*@\circled{1}@*)
    \end{teX}
(b) Determine Attn. Span
\end{minipage}%
\begin{minipage}[t]{0.19\linewidth}
    \captionsetup{name={Fig.}, labelsep=period}
    \begin{teX}
(*@ \hl{10 + 42 * 3=}@*)(*@\circled{1}@*)
(*@ \highlightandunderline{10 + 126=}@*)(*@\circled{2}@*)
    \end{teX}
(c) Generation
\end{minipage}%
\begin{minipage}[t]{0.21\linewidth}
    \captionsetup{labelformat=empty, labelsep=space}
    \begin{teX}
10 + 42 * 3=(*@\circled{1}@*)
(*@\hl{10 + 126=}@*)(*@\circled{2}@*)
    \end{teX}
(d) Determine Attn. Span
\end{minipage}%
\begin{minipage}[t]{0.18\linewidth}
    \captionsetup{labelformat=empty, labelsep=space}
    \begin{teX}
10 + 42 * 3=(*@\circled{1}@*)
(*@\hl{10 + 126=}@*)(*@\circled{2}@*)
(*@\highlightandunderline{136}@*)
    \end{teX}
 (e) Generation
\end{minipage}

\caption{Illustration of Autoregressive Inference with Reduced Attention Span.
In step (b) and (d), the LLM attends to all tokens to select a subset of important tokens for next token prediction.
We highlight the selected tokens in green.
During generation step (c) and (e), the LLM attends only to the selected subset of tokens.}
\label{fig:example}
\end{figure*}

%% file: exp_res.tex
\section{Experimental results}
We present two case studies on our proposed technique.

\textbf{Hardware and software.}
We perform our evaluation on an A100 GPU with 80GB of HBM (High Bandwidth Memory).
With the exception of custom CUDA kernels, we based our code on Hugging Face's implementation of LLaMA models.
As outlined in \Cref{sec:background}, we develop custom CUDA kernels specifically for the decoding phase of inference, due to its unique performance bottlenecks.
Our evaluation mainly targets this part of the model inference.
We always generate tokens using greedy decoding, which picks the most probable next token at each step of generation.
We always pre-allocate the KV-cache to eliminate runtime overhead due to memory allocation.

\paragraph{LORA configuration.}
We fine-tune both sparse and dense attention models with LORA-finetuning using identical hyper-parameters adopted from \citet{liu2023goat} -- we set LORA rank and alpha to $64$, LORA dropout to $0.05$, number of epochs to $1$, learning rate to $3\times10^{-4}$. 
We fine-tune with a causal language-modeling objective.
All models in this section derive from the LLaMA-7B base model.
We fine-tune the K/Q/V and output projection layers of the model.
We always merge LORA weights back into the dense model for inference evaluation for efficiency.

\textbf{Comparison.}
We compare three combination of attention mechanism and kernel implementations.
The first uses dense attention implemented with the standard Huggingface \texttt{transformers} (denoted Dense Attn + Dense Kernel).
This combination represents standard off-the-shelf inference speed.
The second employs sparse attention with a custom CUDA kernel (denoted Sparse Attn + Sparse Kernel) using our proposed technique of sparse attention and the custom CUDA kernel.
The third combines dense attention with a custom CUDA kernel to verify that our efficiency gains are not primarily from optimized attention implementation (denoted as Dense Attn + Sparse Kernel)\footnote{At the time of our writing, Huggingface implementation of LLaMA does not utilize flash attention.}.

\lstdefinestyle{mystyle}{
numbers=left,
numbersep=4pt,                   %
framexleftmargin=6pt,           %
frame=tb,
basicstyle=\footnotesize,
backgroundcolor=\color{white},
escapeinside={(*@}{@*)},
literate={\[0\]}{\colorbox{green}{[0]}}1
{\[1\]}{\colorbox{green}{[1]}}1
{\[2\]}{\colorbox{green}{[2]}}1
}

\begin{figure}[!b]
    \begin{lstlisting}[style=mystyle]
(42 * 56) + (5 * 32) (*@\colorbox{cyan}{[0]}@*)
(*@\colorbox{green}{[-1]}@*) 42 * 56=42 * (50 + 6)=2100 + 252=2352
(*@\colorbox{green}{[-1]}@*) So 42 * 56=2352 (*@\colorbox{cyan}{[1]}@*)
(*@\colorbox{green}{[0,1]}@*) 5 * 32=160 (*@\colorbox{cyan}{[2]}@*)
(*@\colorbox{green}{[0,1,2]}@*) 2512
\end{lstlisting}
    \caption{Example model output for complex arithmetics.
        Numbers on a blue background are anchors.
        Numbers on a green background reference these anchors. 
        [-1] denotes a reference to the previous line.}
    \label{fig:sample-arith-eval-output}
\end{figure}

\begin{figure*}
    \centering
    \includegraphics[width=1.\linewidth,trim={2.5cm .5cm .5cm 0},clip]{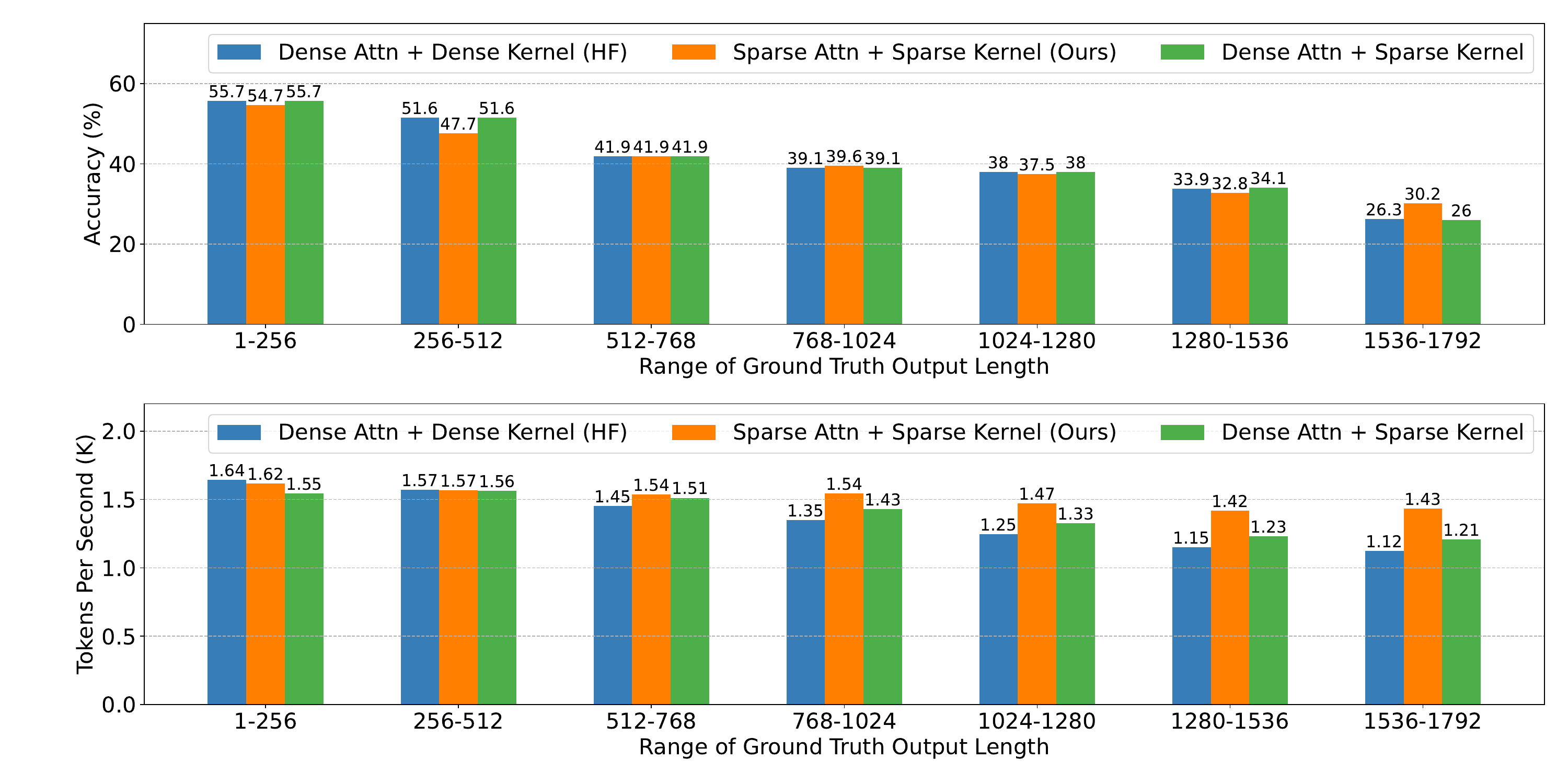}
    \caption{Accuracy and Throughput in Thousands of Tokens/s for Fine-tuned LLaMA 7B Model on Arithmetic Evaluation Task. 
    Our proposed approach (Sparse Attn + Sparse Kernel, Ours) speeds up inference by up to 28\% compared with the Huggingface transformers implementation (Dense Attn + Dense Kernel, HF) without compromising task accuracy.}
    \label{fig:enter-label}
\end{figure*}

\subsection{Arithmetic evaluation task}
\label{subsec:arith-eval}
We apply our proposed technique to evaluation of arithmetic expressions using incremental steps.
This task requires Chain-of-Thought~\citep{DBLP:journals/corr/abs-2201-11903} reasoning to achieve reasonable accuracy.
Due to the lengthy intermediate thought processes involved, we examine the ability of LLMs to focus on relevant previous reasoning steps as they generate solutions in incremental steps.

\textbf{Dataset generation.}
We generate the task dataset using random arithmetic expressions that involve addition, subtraction, multiplication, and integer division.
These expressions have a depth and digit count, both normally distributed with a mean of $5$ and a standard deviation of $2$.

We begin with the complete arithmetic expression and identify all binary operations where both operands are literal numbers.
We then evaluate these operations one by one, generating an evaluation trace for each of them.
If an operation involves multiplication or division with two multi-digit numbers, we decompose the calculation so that each step includes at most one multi-digit number. This method of breaking down intermediate steps for binary arithmetic is adopted from \citet{liu2023goat}.

Throughout the generation process, we maintain a record of logical dependencies between computational steps.
We include corresponding tokens for anchors and references.
We cap the maximum digit count for any single number in the dataset at $10$ digits.
The dataset consists of $60,000$ examples, evenly distributed across expression lengths.

For example, an input instruction to solve $(42 \times 56) + (5 \times 32)$ elicits model output in \Cref{fig:sample-arith-eval-output}:
The model produces an anchor ``[0]" at the end so that it can refer back to the task at hand.
Note that two expressions, $42 \times 56$ and $5 \times 32$ are both ready to evaluate; L2 evaluate the first expression in steps, and L3 summarizes the result of evaluating this first expression.
Both steps begins with special references ``[-1]" to the previous line.
The model finishes L3 with an anchor ``[1]" for future reference.
In L4, the model references the overall task through anchor ``[0]" as well as the already completed sub-expression ``[1]" and determines the next sub-expression to evaluate is $5 \times 32$.
The model computes $5 \times 32$ in L4, along with anchor ``[2]".
Finally, with reference to the overall task ``[0]" and two evaluated sub-expressions ``[1,2]", the model generates the final answer.

\paragraph{Challenges.} This example illustrates that, to solve this task correctly, it is insufficient to simply maintain a constant number of most recent tokens, as the model may refer to tokens from the arbitrarily distant past.

\textbf{Fine-tuning.}
We fine-tune the GOAT model \citep{liu2023goat}, which is a LLaMA-7B model fine-tuned to solve binary arithmetic evaluation.

\textbf{Evaluation metrics.}
We evaluate models using task accuracy and token generation speed.
To calculate task accuracy, we split the model output by space, parse the last term into an integer, and compare it to the ground truth output.
For performance, we report tokens generated per second.

\textbf{Methods.}
We assess our technique by using fine-tuned LLMs to generate step-by-step solutions to evaluate arithmetic expression unseen during fine-tuning.
We generate random arithmetic expressions whose ground truth solutions have up to $N$ tokens.
We divide $N$ evenly into ranges of equal lengths $L$.
For each range, we generate $K$ random expressions for evaluation, and report the corresponding accuracy and throughput for our approach as well as baseline techniques.
We take $N=1536$ because it is reasonably close to the model's context length limit of $2048$ while leaving room for instruction prompt as well as headroom for a slightly more verbose output.
We take $L=256$ and $K=384$.
We set batch size to 48 as it is the largest that avoids out-of-memory errors in all evaluation setups.

\textbf{Results.}
\Cref{fig:enter-label} shows the accuracy and throughput for three combinations of attention mechanism and kernel implementation.
Our proposed sparse attention with a custom CUDA kernel often matches or exceeds the accuracy of dense attention.
Except for output lengths between 256 and 512 tokens, the accuracy of our sparse attention never drops more than 1\% below that of dense attention.
Moreover, our proposed technique sees larger speedup for generating longer sequences.
This is expected, as for longer generations, more opportunities exist to ignore irrelevant tokens in the LLM's context.
Specifically, for output lengths between 1536 and 1792 tokens, our implementation is $27.7$\% faster than the standard Huggingface transformer.

Our custom CUDA kernel is based on FlashAttention \citep{dao2022flashattention}, which may improve performance due to fused attention. 
We thus experiment with executing dense attention with our custom CUDA kernel (shown as ``Dense Attn + Sparse Kernel'' in \Cref{fig:enter-label}).
We configure attention mask to retain all positions in the attention score matrix.
Throughput of this configuration stays within 7\% of the Huggingface transformer implementation in all cases.

\textbf{Conclusion.}
We fine-tune an LLM to evaluate complex arithmetic expressions while minimizing its attention span.
We design and implement a CUDA kernel to take advantage of this reduced attention span and achieved a speedup of up to 28\% without compromising task performance.

\subsection{News article summarization task}
\label{subsec:news-summarization}
We also studied the performance of applying our proposed technique on a summarization task, where the model needs to summarize an article in 3 sentences and only focus on a limited range of article sentences when generating each summarization sentence.

\textbf{Dataset generation.}
Our dataset is generated based on CNN/Daily Mail dataset~\citep{nips15_hermann}, which contains English journal articles from CNN and Daily Mail. For each article, we split it by sentence and prepend a line number to beginning of the sentence and a line break to the end of the sentence. We then pass the article to ChatGPT 3.5 asking it to summarize the article.

The format of output summary consists of 3 sentences where each  sentence is preceded with the range of line numbers from the article indicating its source. 
ChatGPT output is used as the ground truth label for fine-tuning.
 \Cref{sec:prompt} shows how we structure input and output during fine-tuning.

\Cref{fig:sample-summarization} shows an example of the summarization task. The line numbers preceding the article sentences are anchors and the range of contiguous line numbers following the summarization sentences are the references to the anchors. 
During generation, we always attend to the instruction at L1.
The generation of reference attends to all preceeding tokens, whereas the generation of summarization sentence only attends to the range of lines indicated by the line number references, plus the previous summarization sentence.

\begin{figure}
    \begin{lstlisting}[style=mystyle]
Summarize the following article ... 
(*@\colorbox{cyan}{0}@*): Sentence 0 of article.
(*@\colorbox{cyan}{1}@*): Sentence 1 of article.
...
(*@\colorbox{cyan}{30}@*): Sentence 30 of article.
(*@\colorbox{green}{(0-10)}@*) Summarization sentence 1.
(*@\colorbox{green}{(17-19)}@*) Summarization sentence 2.
(*@\colorbox{green}{(25)}@*) Summarization sentence 3.
\end{lstlisting}
    \caption{Example model generation for summarization tasks.
        Numbers on a blue background are anchors.
        Numbers on a green background reference these anchors.}
    \label{fig:sample-summarization}
\end{figure}

\textbf{Fine-tuning.}
We fine-tune both sparse attention and dense attention models using LORA-fine-tuning with a LLaMA Chat 7B model \citep{touvron2023llama}. 

\textbf{Evaluation metrics.} 
We evaluate output summary using ROUGE score~\citep{rouge}. %
We also report number of tokens generated per second to evaluate model throughput.

\begin{table}[!ht]
\centering
\small
\begin{tabular}{lrrrrr}
\toprule
                    Configuration &  Toks/s &  R.1 &  R.2 &  R.L &  R.Lsum \\
\midrule
  DADK Short &      751.8 &   58.6 &   36.5 &   45.9 &      46.8 \\
   DASK Short &      789.2 &   58.5 &   36.3 &   45.8 &      46.6 \\
SASK (ours) Short &      788.8 &   52.2 &   28.0 &   38.5 &      39.9 \\
   DADK Long &      652.9 &   54.1 &   29.3 &   38.6 &      40.1 \\
  DASK Long &      732.5 &   54.0 &   29.2 &   38.6 &      40.0 \\
 SASK (ours) Long &      771.2 &   47.5 &   21.6 &   32.2 &      34.1 \\
\bottomrule
\end{tabular}
\caption{Results of summarization task. DASK = Dense Attention, Sparse Kernel, etc. R. = ROUGE-.}
\label{tbl:summarization-results}
\end{table}

\textbf{Methods.}
We generate summaries of $550$ news articles in our test set.
We divide the test set into two partitions: short and long. 
Articles in the long partition have a total number of tokens between $1536$ and $2048$, including the original article and generated summary, $121$ articles in total.
The rest belongs to the short partition, $408$ articles in total.
We use a batch size of $24$, which is the largest batch size in multiple of $8$, that fits in memory.

\textbf{Results.}
We summarize the accuracy and throughput for each combination of attention sparsity and kernel implementation in \Cref{tbl:summarization-results}.
For each configuration, we specify both the model and kernel configuration. 
We note that whether the dense attention model configuration utilizes the dense or the sparse kernel (with 0\% sparsity) does not, in principle, change its output as these two kernel implementations are semantically equivalent with sparsity set to 0.
Our approach (denoted as SASK) combines \textbf{S}parse \textbf{A}ttention with a custom \textbf{S}parse CUDA \textbf{K}ernel.
Our approach sees the highest throughput on the long partition, beating the Huggingface transformer implementation by $18.2$\% and the baseline dense attention with sparse CUDA kernel by $5.3$\%.

Sparse attention models generally under-perform dense attention models, unlike in the previous case study.
We hypothesize two possible reasons: (1) imprecision of ChatGPT annotation of attention span -- as generated summaries often reference terms outside the annotated range; (2) models are under-fine-tuned on the annotated dataset, for sparse models to adapt to restricted attention span.
While improving annotation quality is beyond our means, we tested hypothesis (2). We increased the fine-tuning epochs for the sparse attention model to 3. This reduced the ROUGE-1 score gap from $6.39$ to $2.2$.

\textbf{Conclusion.}
For summarization task, our proposed approach achieves the highest throughput when summarizing long articles, though at a noticeable cost to output quality.
However, additional fine-tuning can narrow the output quality gap between sparse and dense models.

\section{Additional Studies}

\begin{figure*}[]
    \centering
    \includegraphics[width=1.\linewidth]{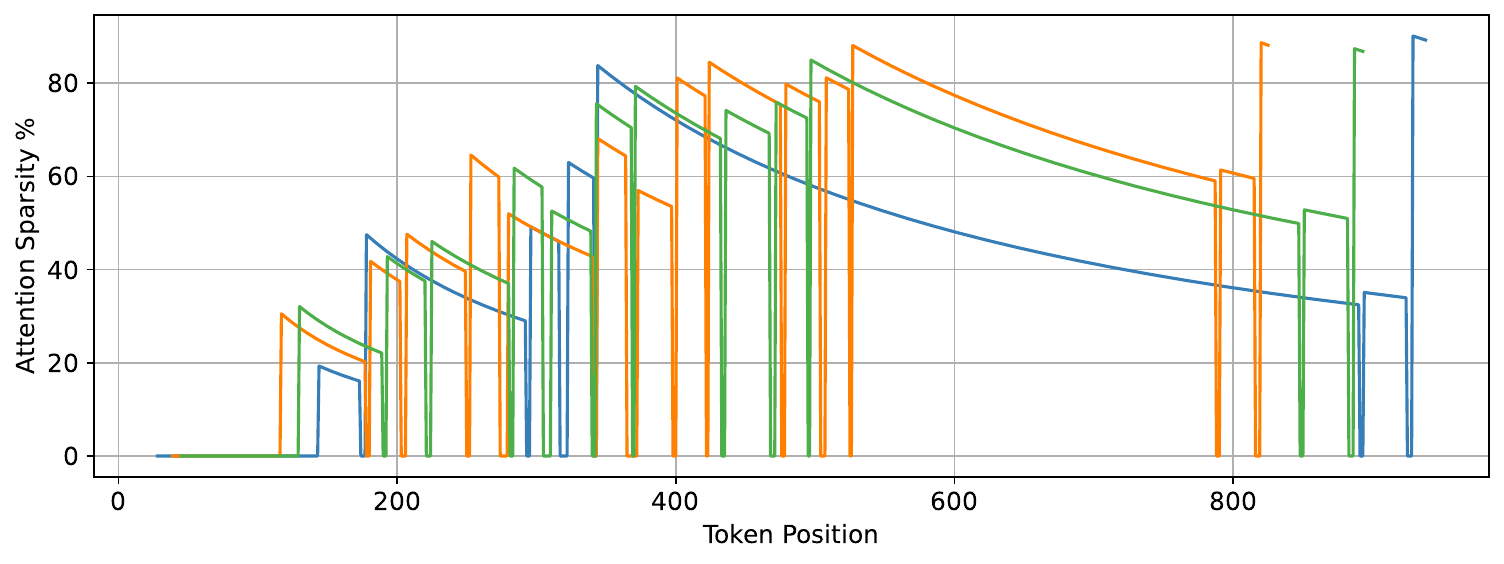}
    \caption{Attention sparsity for generating each token to evaluate complex arithmetic expressions.
    We select expressions from the test dataset.
        The ground truth solution, including the intermediate steps, have length between 768 and 1024 tokens.
        Dips in the attention sparsity to 0 occurs when generating references, which require attention to full context.
        We observe that attention sparsity generally grows as the context length increase, and can reach to more than 80\% near the end of solution generation. }
    \label{fig:sparsity-over-time}
\end{figure*}

\subsection{Achieved sparsity of self-selected attention}

Since we allow the LLM to dynamically select important tokens, the attention sparsity dynamically vary during autoregressive inference.
In this section, we inspect the attention sparsity for each token during autoregressive generation for the arithmetic evaluation task.

\textbf{Method.}
We analyze evaluation data with ground truth output length between $768$ and $1024$.
We select the first three runs and record the achieved attention sparsity for each output token.
The attention sparsity is calculated as the number of ignored tokens divided by the total number of already generated tokens, including the prompt tokens.

\textbf{Results.}
We plot attention sparsity as a function of the position of the generated token during inference in \Cref{fig:sparsity-over-time}.
We observe considerable attention sparsity during autoregressive generation.
Specifically, the average attention sparsity over all token positions and across three runs is $47.3$\%.
The attention sparsity generally increases with growing token positions, because the larger number of previously generated tokens, the more opportunities to selectively ignore irrelevant tokens.
Occasional dips to $0$\% occur when generating references to anchors, which require full-context attention. 

\textbf{Conclusions.}
We observe considerable attention sparsity on average and find that LLMs dynamically adjust the attention sparsity during autoregressive generation.

\subsection{Custom CUDA kernel benchmarking}
\label{custom-cuda-kernel-benchmark}

Our custom CUDA kernels are designed for unstructured sparse attention patterns.
Nevertheless, performance only improves when the attention matrix has clustered nonzero elements.
This is because our implementation computes attention probabilities in blocks:
the implementation computes attention probabilities for blocks with nonzero elements and skips those with only zeros.
Thus, block size matters: too small limits parallelism, too large increases computation.

To understand this tradeoff quantitatively, we examine our custom CUDA kernels in this section, and investigate the speedup achievable under controlled conditions.

\begin{figure}[b!]
    \centering
    \includegraphics[width=1.2\linewidth, trim={1.5cm .5cm 0 2cm},clip]{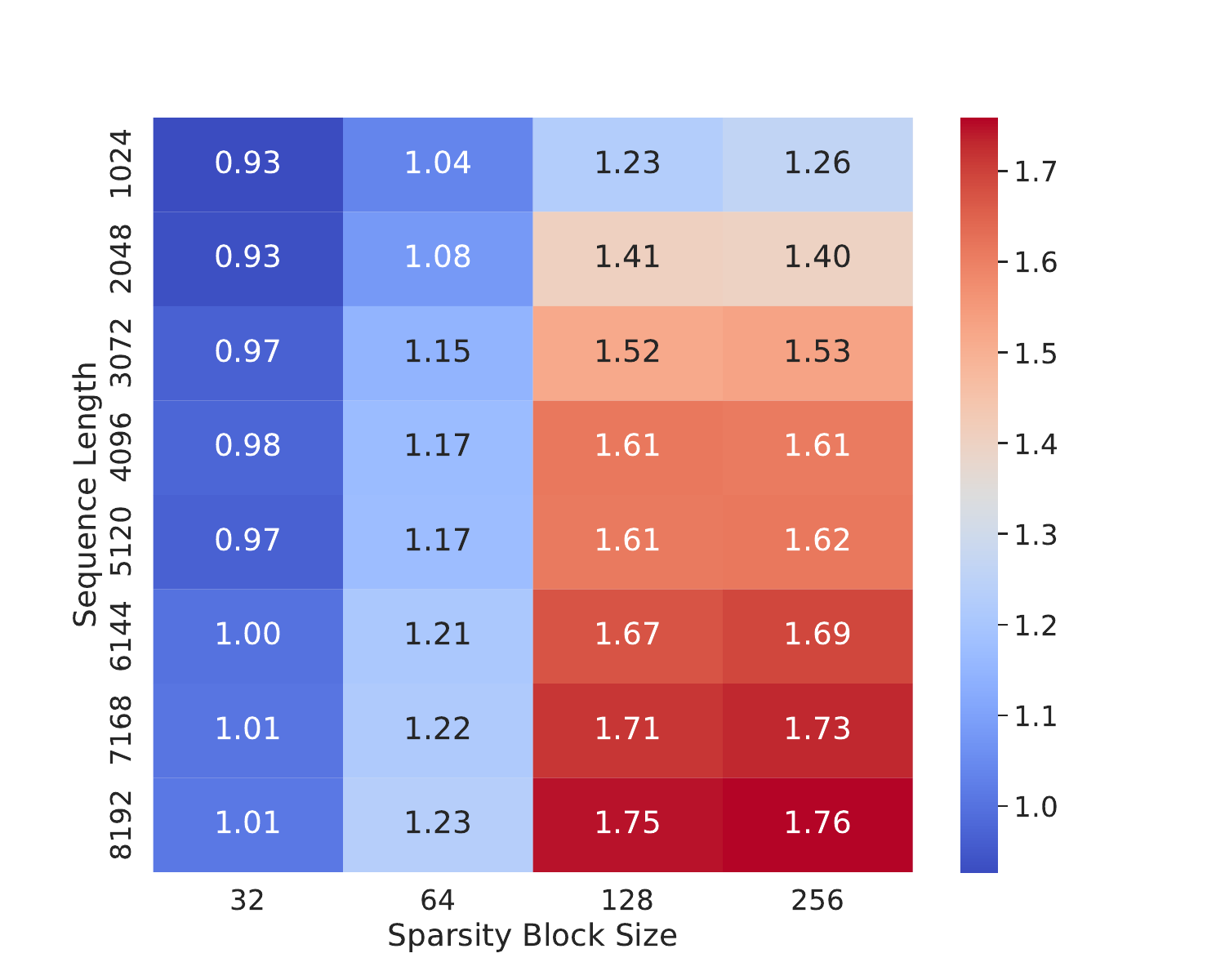}
    \caption{Speedup under controlled conditions.
    Our custom CUDA kernel speeds up inference when ignored context occur in clusters of $64$ or more tokens.}
    \label{fig:speedup-results}
\end{figure}

\textbf{Method.}
We simulate generating a new token with varying lengths of preceding tokens, among which $50$\% are selectively ignored.
Specifically, we randomly generate binary attention masks with a $50$\% sparsity rate.
Within these masks, a value of zero indicates a position in the attention matrix to skip
                 and a value of one indicates a position in the attention matrix to compute.
We fill the mask with blocks of 1s, and we refer to this block size as \textit{sparsity block size}.
We use sparsity block size to control the granularity of sparsity patterns.
Importantly, this sparsity block size should not be confused with the block size used in our custom CUDA kernels.
Regardless of its block size setting, our CUDA kernel can compute attention probabilities for masks generated with any sparsity block size; the distinction affects performance only.
Specifically, a mismatch between these two block sizes can add computational overhead.
We benchmark our CUDA kernel with a fixed batch size of $64$, \footnote{We searched for the maximum batch size with an increment of 16, this is the largest that does not result in compilation error or out-of-memory error.}
varying number of pre-existing tokens from 1024 to 8096 and varying sparsity block size among \{$64$, $128$, $256$\}.
We show speedup against the standard Huggingface implementation of self-attention (with pre-allocated KV-cache).

\textbf{Results.}
As shown in \Cref{fig:speedup-results}, at 50\% sparsity, our custom CUDA kernel delivers a (geometric) average speedup of $1.29\times$ over the standard Huggingface transformers implementation.
The speedup generally increases with sparsity block sizes; the ability to skip over large blocks of zeros is critical for achieving speedup.
The geometric average speedup for sparsity block size $32$, $64$, $128$ and $256$ are $0.97\times$, $1.2\times$, $1.6\times$, $1.6\times$ respectively.
However, our implementation also have small but noticeable overhead due to the additional overhead related with handling the attention mask.
Specifically, with sequence length of $1024$ and sparsity block size of $32$, our implementation is slower than baseline by $7$\%.

\textbf{Conclusions.}
Our custom CUDA kernel is effective at accelerating autoregressive inference when the ignored context occur in blocks of 64 or more tokens.

%% file: related.tex
\section{Related Work}

\textbf{Meta-learning.}
Meta-learning optimizes learning and prediction algorithms, often enhancing system efficiency. 
For example, \citet{han2015learning, li2020dhp} learned neural network connections and weights. 
They evaluate the importance of each connection by its magnitude and pruned unimportant ones, reducing both storage and computation during prediction. \citet{jaderberg2017decoupled, czarnecki2017understanding} used neural networks to predict future gradients, allowing layers to update independently and thus increasing parallelism. 
A substantial body of work \citep{Almeida1999ParameterAI, MartnezRubio2018ConvergenceAO, luketina2016scalable, chandra2022gradient} optimizes hyperparameters, such as learning rate, using gradient-based techniques to minimize tuning. 
We also optimize, using LLMs themselves, hyperparameters controlling the attention span of large language models during inference; unlike traditional, static hyperparameters such as the window size in sliding window attention, we train LLMs to predict and minimize its own attention span.

\textbf{Sparse attention.}
Sparse attention has a history nearly as long as the transformer itself; early work leveraged static structured sparse attention to learn long sequences with limited hardware resources, e.g.~\citet{child2019generating, beltagy2020longformer, zaheer2021big} and numerous subsequent studies, training models with sparse attention.  
Recent studies proposed attention architectures with context-sparsification mechanisms that are learnable (e.g.~\citealt{tay2020sparse,rao2021dynamicvit, mohtashami2023landmark}), training a dynamic KV-cache manager effective at inference-time.  
More related to our work is recent work that employs post-training attention sparsification in a context-dependent manner (e.g.~\citealt{zhang2023h2o,ge2023model,xiao2023efficient}); in these techniques, specifically, a \emph{contextual attention sparsifier} is responsible for administering an \emph{eviction policy} of tokens from the context, which was specifically co-designed for the model and task based on statistics of the attention scores, empirically measured from \emph{BERTological introspections}. 
In this work, we aim at the same upshot while attempting to use an LLM to determine the eviction policy implicitly and adaptively.